\documentclass{article} 
\usepackage{iclr2023_conference,times}

\usepackage{amsmath,amsfonts,bm}

\def\eqref#1{equation~\ref{#1}}

\def\1{\bm{1}}

\DeclareMathAlphabet{\mathsfit}{\encodingdefault}{\sfdefault}{m}{sl}
\SetMathAlphabet{\mathsfit}{bold}{\encodingdefault}{\sfdefault}{bx}{n}

\usepackage{hyperref}
\usepackage{url}

\usepackage{mathtools}
\usepackage{amssymb}
\usepackage{pifont}
\usepackage{subcaption}
\usepackage{booktabs}
\usepackage{multirow}
\usepackage{array} 
\usepackage{algorithm}
\usepackage{algpseudocode}
\usepackage{xcolor, colortbl}
\usepackage{graphicx}
\usepackage{tikz}
\usetikzlibrary{shapes}
\usepackage{soul}
\usepackage{multicol}

\newcommand{\xmark}{\ding{55}}%

\title{REST: REtrieve \& Self-Train for generative action recognition}

\iclrfinalcopy

\author{Adrian Bulat  \\
Samsung AI Cambridge\\
\texttt{adrian@adrianbulat.com}
\And
Enrique Sanchez \\
Samsung AI Cambridge \\
\texttt{e.lozano@samsung.com} \\
\AND
Brais Matinez \\
Samsung AI Cambridge \\
\texttt{brais.a@samsung.com}
\And
Georgios Tzimiropoulos \\
Samsung AI Cambridge \\
Queen Mary University of London \\
\texttt{g.tzimiropoulos@qmul.ac.uk}
}

\begin{document}

\maketitle

\begin{abstract}
This work is on training a generative action/video recognition model whose output is a free-form action-specific caption describing the video (rather than an action class label). A generative approach has practical advantages like producing more fine-grained and human-readable output, and being naturally open-world. To this end, we propose  to adapt a pre-trained \textit{generative} Vision \& Language (V\&L) Foundation Model for video/action recognition. While recently there have been a few attempts to adapt V\&L models trained with contrastive learning (e.g. CLIP) for video/action, to the best of our knowledge, we propose the very first method that sets outs to accomplish this goal for a generative model. We firstly show that direct fine-tuning of a generative model to produce action classes suffers from severe overfitting. To alleviate this, we introduce REST, a training framework consisting of two key components: an unsupervised method for adapting the generative model to action/video by means of pseudo-caption generation and \textit{Self-training}, i.e. without using any action-specific labels; (b) a \textit{Retrieval} approach based on CLIP for discovering a diverse set of pseudo-captions for each video to train the model. Importantly, we show that both components are necessary to obtain high accuracy. We evaluate REST on the problem of zero-shot action recognition where we show that our approach is very competitive when compared to contrastive learning-based methods. Code will be made \href{https://www.adrianbulat.com/generative-action-recognition}{available}.
\end{abstract}

\section{Introduction}

Large-scale pre-training of Foundation Models (FM) for Vision-Language (V\&L) understanding~\citep{radford2021learning,jia2021scaling, yuan2021florence, yu2022coca, alayrac2022flamingo, li2022blip, wang2022git} has recently revolutionized several multi-modal understanding tasks including image captioning, VQA, text-based retrieval, and visual reasoning. Moreover, recently, such models, and especially CLIP~\citep{radford2021learning}, have been also applied to more traditional visual tasks including object detection~\citep{gu2021open, du2022learning, minderer2022simple} and video/action recognition~\citep{wang2021actionclip, ju2022prompting, castro2022fitclip, lin2022frozen}. This work is on adapting a V\&L FM for video/action recognition but in contrast to all prior work that use a V\&L model trained with contrastive learning, i.e. CLIP~\citep{wang2021actionclip, ju2022prompting, castro2022fitclip, lin2022frozen}, our focus is on adapting a \textit{generative} V\&L model, and specifically BLIP~\citep{li2022blip}, to the video domain with the goal of producing an action-specific caption in an autoregressive manner. Besides this being a challenging research goal on its own, a generative approach has practical advantages including producing more fine-grained (compared to an action class label) and more interpretable/human-readable output, and being naturally more open-world (compared to contrastive learning based approaches) in a sense that it does not require a priori definition of the action classes.    

In prior work~\citep{wang2021actionclip, ju2022prompting, castro2022fitclip, lin2022frozen}, CLIP adaptation to action/video recognition was seamlessly done using the standard contrastive loss and hand-engineered or learned prompts encoding action class information. In contrast, as Table~\ref{tab:zero_shot_k600_finegrained} shows the direct end-to-end fine-tuning of the model on class names suffers from severe overfitting and poor open-world (i.e. zero-shot) generalizability, a fundamental property of V\&L models that we wish not to sacrifice when adapting the model to the video domain. 

\begin{table}[ht]
    \centering
        \caption{Zero-shot generalization of our Generative Action Recognition (GAR) model trained with action class labels vs action-specific pseudo-caption as proposed in REST. We report generalized zero-shot classification results on Kinetics-220 (1-vs-620 setting; see Sec.~\ref{sec:results} for more details). We observe that training with class labels results in very poor zero-shot generalization. Upon inspecting the model's predictions, we observe that it does indeed restrict itself to the class names seen on the training set. Some examples of failure cases (expected - predicted): ``massaging neck'' - ``massaging head'', ``playing ocarina'' - ``playing harmonic'', ``separating eggs'' - ``scrambling eggs'', ``putting on lipstick'' - ``sticking tongue out'' etc.}
    \label{tab:zero_shot_k600_finegrained}
    \begin{tabular}{lccc}
    \toprule
         Method & pseudo-captions & {Top-1} & {Top-5} \\
         \midrule
         GAR + class labels & \xmark & $0.76 \pm 0.4$ & $37.8\pm1.0$ \\
         GAR + REST (this paper) & \checkmark & $\mathbf{29.51\pm0.71}$ & $\mathbf{56.12\pm0.37}$ \\
         \bottomrule
    \end{tabular}
    \vspace*{-0.5cm}
\end{table}

To alleviate the aforementioned overfitting problem, we propose to entirely drop action labels during training and adopt an unsupervised adaptation approach. Specifically, starting from an image-based generative model (specifically BLIP~\citep{li2022blip}), we describe a self-training procedure where the model is iteratively trained on action data, without any labels, using an auto-regressive objective where the training pseudo-captions are produced by the model itself. Not only the model, in this case, is able to generalize much better to unseen classes but also such an approach is label-free making it potentially suitable to be applied to any large-scale video dataset. However, our findings suggest that self-training solely is not sufficient to train a highly accurate model as the generated pseudo-captions are lacking diversity~\footnote{Some sort of the so-called confirmation bias~\citep{arazo2020pseudo}}. To alleviate this, we further propose a retrieval approach which is integrated into self-training where every few epochs a CLIP model is used to retrieve, for each training video, additional pseudo-captions from similar videos (from the same dataset). These pseudo-captions are then used to enhance learning by means of increasing the diversity of the training data. Critically, we show that our retrieval approach drastically increases the accuracy and the quality of the trained models. In summary, our \textbf{main contributions} are:

\begin{itemize}
    \item To the best of our knowledge, we propose the very first method for adapting a \textit{generative} V\&L FM for open-vocabulary action/video recognition.  
    \item 
    We introduce REtrieve \& Self-Train (REST), a training framework consisting of two key components; (a) iterative \textit{Self-training} without using any action-specific labels, and (b) CLIP-based \textit{Retrieval} for discovering a diverse set of pseudo-captions to train the model. We show that both components are necessary to train a high quality model.
    \item
    We evaluate REST on zero-shot action recognition where we show that our approach is very competitive with respect to CLIP-based methods, matching and/or surpassing the state-of-the-art on standard evaluation benchmarks.
\end{itemize}

\section{Related work}

\noindent \textbf{Vision-Language Foundation Models:} Following CLIP and Align~\citep{radford2021learning, jia2021scaling}, a number of methods were proposed that train Vision \& Language (V\&L) Foundation Models with contrastive learning~\citep{li2021supervision,yao2021filip, yu2022coca, zhai2022lit}. More recently, a few attempts were made towards training generative (V\&L) models that produce as output a caption describing the input image in a auto-regressive manner~\citep{tsimpoukelli2021multimodal,li2022blip, alayrac2022flamingo, wang2022git}. From these methods, we used BLIP~\citep{li2022blip} as the start point of our framework, the goal of which is to produce a generative action recognition model that operates in the video domain in an unsupervised manner. BLIP was chosen on the basis that publicly available models are provided but, in practice, any available image-based generative model can be incorporated into our training pipeline.

\noindent \textbf{Unsupervised \& Semi-supervised Image Captioning:} There are only very few methods that have attempted to train an image captioning model without full supervision. ~\citep{chen2017show} propose a method that transfers a COCO model to other domains by means of adversarial training using unpaired data in the target domain.
Similar in spirit approaches focusing on training an LSTM discriminator to distinguish real from generated captions were proposed in~\citet{feng2019unsupervised}, \\\citet{laina2019towards,zhou2021triple}. Moreover, semi-supervised approaches include~\citet{kim2019image} which combines paired and unpaired data with adversarial training, and~\citet{chen2021self} which
performs iterative self-training using a mean teacher consisting of an ensemble of independently trained models. Compared to~\citet{chen2021self}, which is the closest work to ours from the above, our method (a) does not use a mean teacher but critically a \textit{retrieval} component which is shown to greatly improve the generated captions, and (b) focuses on image-to-action/video captioning (rather than on image-to-image) which is significantly more challenging.

\noindent \textbf{V\&L Foundation Models for Action/Video Recognition:} Following the development of CLIP, a number of very recent works have attempted to adapt it to the video domain. X-CLIP~\citep{lin2022frozen} trains a lightweight transformer on top of CLIP image features for spatiotemporal fusion.~\citet{ju2022prompting} uses soft prompt learning to adapt CLIP to the video domain, while~\citet{wang2021actionclip} performs standard end-to-end finetuning. Starting from CLIP, FitCLIP~\citep{castro2022fitclip} proposes a teacher-student approach based on a small video-text labelled dataset and pseudo-labels generated on a large unlabelled dataset. Notably, their method uses an ensemble largely relying on the original CLIP model to produce high accuracy. The aforementioned works use 
contrastive learning and labelled data for CLIP-to-video adaptation. In contrast, our work is the first, to our knowledge, that attempts to adapt a generative 
V\&L model to the action/video domain, which on its own poses significant difficulties, and notably, without using any action-specific labels.

\noindent \textbf{Video Captioning:} A large body of methods are trained in a fully supervised manner using caption annotations (which are costly), see for example~\citet{pan2020spatio,tang2021clipplus,tang2021clip4caption, lin2021end-to-end, liu2022video}. In contrast, REST does not use any human annotations to train the models.

\noindent \textbf{Zero-shot Action Recognition:} Most works on zero-shot action recognition build on learning attribute-based \textit{semantic embeddings} from video features in order to make them be close to the word embedding of the class names~\citep{zhu2018towards, brattoli2020rethinking, bretti2021zero,mettes2022universal,estevam2022global, pu2022alignment,luo2022disentangled}, or use a text encoder which is learned or updated as part of the training~\citep{ni2022expanding,ge2022bridging,qian2022multimodal,lin2022cross,qian2022multimodal}. Most recent approaches using word embeddings focus on the \textit{alignment} problem i.e. learning visual representations that match the corresponding class semantic embeddings but also generalize to unseen class embeddings~\citep{pu2022alignment,luo2022disentangled}. All the aforementioned methods above operate in a discriminative setting where the class embeddings need to be computed. In contrast, our method is a generative one, being able to generate action-specific captions in an autoregressive manner without having to manually pre-define the classes of interest.

\section{Method}\label{sec:method}

\subsection{Overview of REtrieve \& Self-Train (REST)}

We are given an action/video recognition dataset $D$ consisting of $N$ video clips $v_i,\;i=1,\dots,N$ \textit{without} the class labels. We construct a generative action recognition model consisting of a video encoder $g_v(.)$ and an autoregressive text decoder $g_t(.)$, both instantiated from a pre-trained generative V\&L model, specifically BLIP ~\citep{li2022blip} (see Sec.~\ref{ssec:GAR}). Our objective is to iteratively train the generative action recognition model on pseudo-captions produced by the model itself (self-training) using a standard language modeling loss (see Sec.~\ref{ssec:LM}). The self-training process is greatly enhanced by increasing the diversity of the pseudo-captions by using a retrieval module based on CLIP (see Sec.~\ref{ssec:CLIP}). The integrated retrieve \& self-train framework is described in Sec.~\ref{ssec:REST}.

\begin{figure}[ht]
    \centering
    \includegraphics[height=5.0cm,trim={0.5cm 12.0cm 9.0cm 1.5cm},clip]{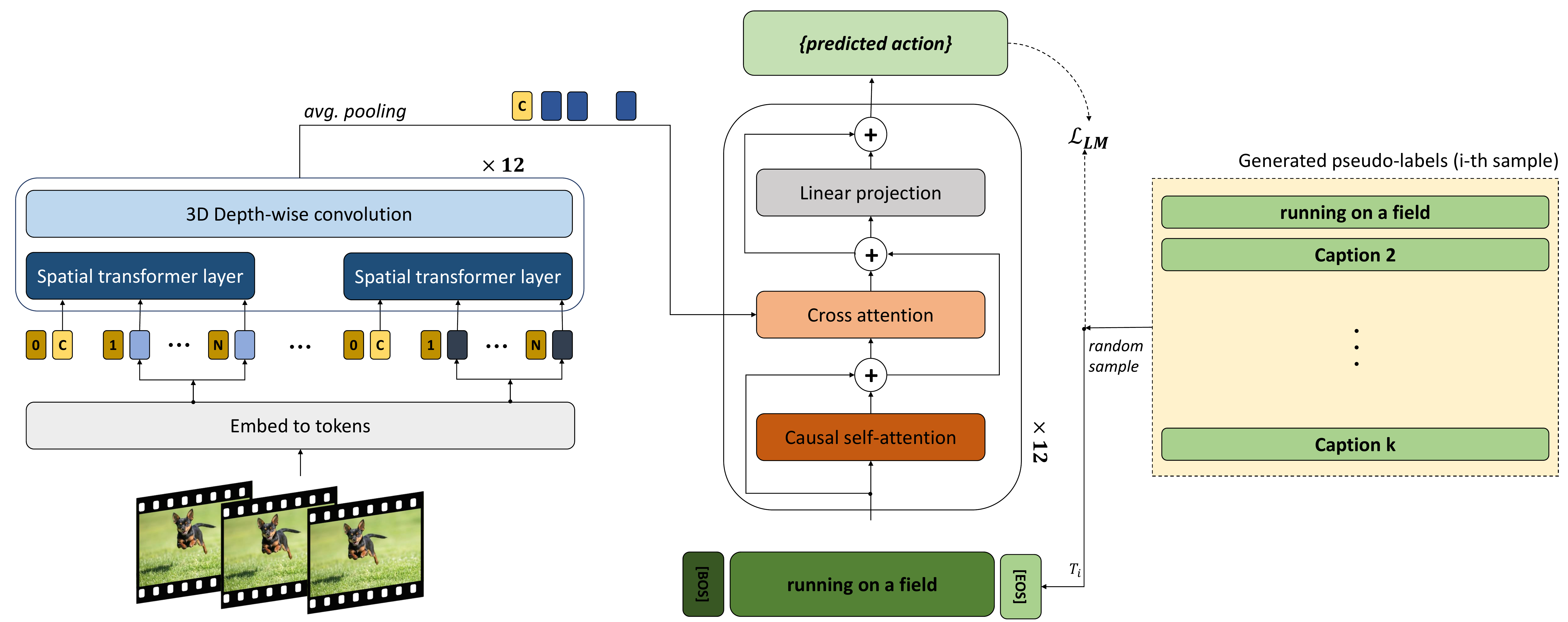}
    \caption{\textbf{Generative Action Recognition network.} We largely maintain BLIP's generative architecture~\citep{li2022blip} adding only a lightweight temporal adapter, in a form of a 3D depth-wise convolution, to combine information across frames.}
    \label{fig:action_recognition_network}
    \vspace*{-0.3cm}
\end{figure}

\subsection{Generative action recognition model} \label{ssec:GAR}

Both the video encoder and the autoregressive text decoder are instantiated from a BLIP model~\citep{li2022blip} pretrained on a large scale image-text dataset. We largely maintain BLIP's generative architecture by only slightly modifying its visual transformer to combine temporal information across frames using a temporal adapter. Fig.~\ref{fig:action_recognition_network} depicts our generative action recognition network.  

The \textbf{text decoder} $g_t(.)$ consists of 12 transformer layers, each performing causal attention on the text tokens and cross-attention between the text and the visual tokens produced by the video encoder. 

The \textbf{video encoder} $g_v(.)$ is constructed by inserting a temporal adapter in-between the layers of BLIP's spatial visual transformer (a ViT~\citep{dosovitskiy2020image}). The visual transformer layers continue to operate independently on each frame (i.e. spatial attention only) while the adapter will pool and combine information across frames, handling the temporal modeling aspect. Specifically, let $\mathbf{Z}^l\in\mathbb{R}^{T\times H \times W \times d }$ be the output of the spatial transformer's layer $l$ (the class token $\mathbf{Z}_{cls}^l$ is excluded). The adapter operates on $\mathbf{Z}^l$ as follows:
\begin{equation}
    \mathbf{Z'}^l = \mathbf{Z}^l + a_d(\mathbf{Z}^l),
\end{equation}
where $a_d(.)$ takes the form of a $3\times 3\times 3$ 3D depth-wise convolution. The adapter layer is initialized such that the temporal residual, introduced by it, will be of low magnitude initially, closely aligning the network's output early on to that of the pre-trained image model. This helps stabilizing the training by avoiding large magnitude gradients caused by incoherent predictions.

Finally, $\mathbf{Z'}^l$, along with the class tokens $\mathbf{Z}_{cls}^l$, are averaged over $T$, to compute video feature $\mathbf{F}^l\in\mathbb{R}^{(H \times W +1) \times d }$ that interacts with the text decoder.

\subsection{Language modelling loss} \label{ssec:LM}

The language modelling objective $\mathcal{L}_{LM}$ is the negative log-likelihood of the self-generated pseudo-captions, and it is used as the main loss to update our model $g_v(.)$ and $g_t(.)$ To compute the loss, we compute the visual tokens $\textbf{F} \in \mathbb{R}^{(H\times W +1 )\times d} = g_v(\mathbf{X})$ from the video frames $\mathbf{X}$, and we are given a pseudo-caption $\tilde{w}$. As we shall see in Sec.~\ref{ssec:REST}, $\tilde{w}$ is sampled from a set of cached pseudo-captions initially generated by BLIP and then updated by $g_t(.)$ over the course of training. We compute the text tokens as $y_{\tilde{w}} = \phi(\tilde{w}) \in \mathbb{N}^M$, with $\phi$ being a tokenizer function that maps (sub-)words into the one-hot vectors spanning the vocabulary size. Following standard practices in image captioning, we prepend to $y_{\tilde{w}}$ a $y_0 = [\texttt{BOS}]$ token as well as a prompt $y_p = \phi(``\texttt{A video of }") \in \mathbb{N}^P$. The input to the model's decoder $g_t(.)$ is set as $y = [y_0, y_p, y_{\tilde{w}} ] \in \mathbb{N}^{1+P+M}$. The text decoder applies left-to-right masked attention (i.e. causal attention) and produces an output $o = [o_i]_{i=1}^{1+P+M}$ with $o_i = g_t(y_{i'<i} | \mathbf{F})$. The language modelling loss is then computed using standard cross-entropy (CE):
\begin{equation}
\mathcal{L}_{LM}(y_{\tilde{w}}) = \sum_{i=1+P}^{1+P+M} CE(y_i, o_i) \label{eq:LM}.
\end{equation}

\subsection{Video-Video \& Video-Text Retrieval} \label{ssec:CLIP}

A key component of REST is the integration of a retrieval module into the self-training process. Our framework uses video-video and video-text retrieval modules both instantiated from a pretrained CLIP model~\citep{radford2021learning} that remains frozen over the training process. We selected CLIP due to its accuracy and strong generalizability properties~\citep{radford2021learning,alayrac2022flamingo}. 

Given a video $\mathbf{X}$ and pseudo-caption $w$, we use the CLIP image $g^C_I(.)$ and text $g^C_T(.)$ encoders to compute video $\mathbf{f}_C = \sum_t g^C_I(\mathbf{X}_t)\in\mathbb{R}^d$ and text $\mathbf{t}_C=g^C_T(w)\in\mathbb{R}^d$ features, respectively. For a given pair of videos $i,j$, and a given pair of video-caption $i,j$, the video-video $s_{vv}$ and video-text $s_{vt}$ similarities are computed as: 
\vspace*{-0.7cm}
\begin{center}
\begin{tabular}{p{4cm}p{4cm}}
\begin{equation}
    s^{i,j}_{vv} = \mathbf{f}^i_C\cdot \mathbf{f}^j_C \label{eq:Svv} 
    \end{equation}  &
    \begin{equation}
    s^{i,j}_{vt} = \mathbf{f}^i_C \cdot \mathbf{t}^j_C, \label{eq:Svt}
    \end{equation}
    \end{tabular}
\end{center}
\vspace*{-0.5cm}
where we use the subscript $C$ to refer to features computed by the frozen CLIP model. 

\subsection{REtrieve \& Self-Train (REST)}~\label{ssec:REST}
This section describes our framework for training the model of Sec.~\ref{ssec:GAR} without using any human-annotated action classes/captions. Instead, the model is trained on pseudo-captions generated by the model itself in a self-training manner. We also introduce 
a retrieval module based on CLIP for increasing the diversity of the pseudo-captions used to supervise the training for each input video, which is shown to greatly enhance the learning process. REST is summarised in Algorithm~\ref{alg:method}.

\begin{algorithm}
\caption{REST Training}\label{alg:method}
\begin{algorithmic}[1]
\Require $\{v_i\}, \; i=1:N$ clips, pre-trained CLIP and BLIP models.
\State Compute $\tilde{W}_i = [\tilde{w}_{i;t} = g_v^{BLIP}(\mathbf{X}_{i;t})]$  \Comment{$\;t=1,\dots,T \;,  \; i=1,\dots,N$}
\State Compute $\mathbf{f}^i_C, \mathbf{t}^i_C = g_{I,T}^C(\mathbf{X}_i ; \tilde{W}_i)$ (Sec.~\ref{ssec:CLIP}) \Comment{$\, i=1,\dots,N$}
\State Compute $s^{i,j}_{vv}$ (Eq.~\ref{eq:Svv}) and $s^{i,j}_{vt}$ (Eq.~\ref{eq:Svt})              \Comment{$\, i=1,\dots,N, \; j=1,\dots,N$}
\State Update $\tilde{W}_i$ (Eqs.~\ref{eq:omega}-\ref{eq:update_wi}) \Comment{Retrieve top-$K$ captions from top-$H$ similar videos}    

\While{training}
    \For{$R$ epochs}
        \State Sample batch $v_i$ and $\tilde{w}_i \in \tilde{W}_i$
        \State Update $g_t$ and $g_v$ using Eq.~\ref{eq:LM}
    \EndFor
    \State Compute $\tilde{W}_i \leftarrow \tilde{W}_i \cup [\tilde{w}_{i;new} = g_t(g_v(\mathbf{X}_{i}))]$  \Comment{$\; i=1,\dots,N$}
    \State Update $\mathbf{t}^i_C = g_T^C(\tilde{W}_i)$  \Comment{$\, i=1,\dots,N$}
    \State Compute $s^{i,j}_{vv}$ (Eq.~\ref{eq:Svv}) and $s^{i,j}_{vt}$ (Eq.~\ref{eq:Svt})              \Comment{$\, i=1,\dots,N, \; j=1,\dots,N$}

    \State Update $\tilde{W}_i$ (Eqs.~\ref{eq:omega}-\ref{eq:update_wi}) \Comment{Retrieve top-$K$ captions from top-$H$ similar videos}   
\EndWhile
\end{algorithmic}
\end{algorithm}

\noindent \textbf{Self-train:} The process is iterative and, at each training iteration, video $v_i$ maintains a list of $K$ associated pseudo-captions $\tilde{W}_i = [\tilde{w}_{i,k}],\;k=1,\dots,K$ describing its content. From this list, a pseudo-caption is randomly sampled and used as supervisory signal for $v_i$ to train the model using Eq.~\ref{eq:LM}. After training for $R$ epochs, the model produces a new pseudo-caption for $v_i$ denoted as $\tilde{w}_{i,new}$. This is added to the existing list resulting in $K+1$ pseudo-captions $\tilde{W}_i \leftarrow \tilde{W}_i \cup \tilde{w}_{i,new}$. 

\noindent \textbf{Retrieve:} The retrieval module is used to update the pseudo-caption list for each video $v_i$ over the course of self-training. Specifically, for each $v_i$, we use the video-video similarity of Eq.~\ref{eq:Svv} to compute $\mathbf{s}^i_{vv}=s^{i,j}_{vv}, \;j=1,\dots,N$, and then retrieve the corresponding $K+1$ captions associated to the $H$ most similar videos to $v_i$, creating a large list of $H(K+1)$ captions as:
\begin{equation}
\Omega_i =  \bigcup_{j \in \text{top-}H} \tilde{W}_j      \label{eq:omega}.
\end{equation}
We then compute the CLIP-based text embeddings $\mathbf{t}^j_C = g_T^C(\tilde{w}_j)$ for $\tilde{w}_j \in \Omega_i$, and the $H(K+1)$ video-text similarity scores $s^{i,j}_{vt} = \mathbf{f}^i_C \cdot \mathbf{t}^j_C$. Finally, we update $\tilde{W}_i$ for video $v_i$ by keeping the top-$K$ most relevant captions from $\tilde{W}_i \cup\Omega_i$, i.e. we update the list of captions $\tilde{W}_i$ for video $v_i$ as: 
\begin{equation}
     \tilde{W}_i \leftarrow \underset{i \in \text{top-}K}{\bigcup} \left[\tilde{w}_{i} \in \left[\tilde{W}_i \cup\Omega_i \right] \right].
    \label{eq:update_wi}
\end{equation}

\noindent \textbf{Initialization:} At the beginning of the training process, the videos $v_i,\;i=1,\dots,N$ are populated with pseudo-captions by an off-the-shelf captioning model. To this end, we use BLIP's text decoder to produce a caption $\tilde{w}_{i,t}$ for a set of $T$ video frames $t = 1,\dots, T$. To encourage the generation of action-specific outputs, we use a set of manual prompts, such as ``\texttt{a video of}'', ``\texttt{a person is}'' or ``\texttt{someone is}'', to initialize the text decoder's output. 
\noindent \textbf{Training efficiency:} Besides training the model with the language modelling loss, the above self-training framework includes a retrieval step which can potentially render the training process slow. However, note that the video-video scores $\mathbf{s}^i_{vv}$ for each video $v_i$ are computed using a frozen CLIP model, and, hence, all these scores can be pre-computed and re-used over the course of training. Moreover, the video feature used to compute the video-text scores $\mathbf{s}^i_{vt}$ can also be pre-computed for all videos, and only the text features corresponding to newly produced captions need be computed during training. Hence, $\mathbf{s}^i_{vt}$ can be also efficiently computed during training.

\section{Evaluation of Generative Action Recognition}

While evaluating standard classification models is trivial, this is not the case for our generative action recognition model, given that its output is free unconstrained text. Direct character-by-character assessment is complicated as (a) there is more than one action caption that could describe the video, and (b) the same action can be expressed in multiple ways. Assessing the quality of generated text is an open research question, and various metrics such as CIDEr~\citep{vedantam2015cider}, BLEU~\citep{papineni2002bleu} and ROUGE~\citep{lin2004rouge} have been proposed to evaluate the correctness of the generated captions from the perspective of human judgment. However, such metrics are hard to correlate with the typical accuracy score expected in a classification problem and tend to penalize predictions outside the expected vocabulary disproportionately. 

To alleviate this, we introduce a novel \textbf{CLIP-based Text Accuracy Metric} (CLIP-TAM), that capitalizes on the capacity of the CLIP text encoder to produce semantically distinctive features, ignoring elements of grammar completeness such as pronouns or adverbs, and being also invariant to permutations in the position of the class names. Given a set of class names $\mathcal{C} = \{\gamma_1, \dots \gamma_{|\mathcal{C}|}\}$, we compute the class embeddings using the CLIP text encoder $\mathbf{w}_c = g_T^C(\gamma_c)$, compute the CLIP embedding for a caption $\tilde{w}_i$ generated by our model as $\mathbf{t}_C^i = g_T^C(\tilde{w}_i)$, and select the target class from $\mathcal{C}$ as $\hat{c} = \arg \max_c \mathbf{w}_c \cdot \mathbf{t}_C^i$. By associating a class to the predicted caption, we can directly compute an accuracy score, reducing the problem to a classic closed-set classification one. 

\textbf{User study:} We evaluate the capacity of CLIP-TAM to correlate with human judgment by conducting a user study on 1,000 videos randomly selected from Kinetics-600. Following~\citet{levinboim2019quality}, we formulate the correctness of a video-caption pair as the binomial probability $\hat{p} = P(CORRECT|video,caption)$ that can be estimated from the Bernoulli process. Each trial corresponds to a different human evaluator. The evaluators are shown an input video, which can be visualised multiple times, and are concomitantly asked the following question: \textit{Does the text describe the action shown in the video?}. The raters are requested to choose between \textit{yes} or \textit{no}. To avoid inducing any bias, the interpretation of what represents a ``correct'' caption is left to the human evaluators to interpret and decide. While this could lead to unstable and inconsistent ratings,~\citet{levinboim2019quality} showed that with sufficient annotators (8-10) the results become stable and reproducible. 
The final accuracy score is produced by taking the average of the binary annotations of each annotator on a per-sample basis and thresholding it to 0.5.

\textbf{Conclusions:} The results of the above experiment showed that human annotators agree with the CLIP-TAM metric in 75.67\% of the cases further validating the correctness of the latter to act as a proxy for classification. The accuracy measured by CLIP-TAM on the Kinetics-600 subset was 71.0\% while the annotators' accuracy was 73.80\%. 

\section{Results}\label{sec:results}
\subsection{Experimental setting}\label{ssec:experimental setting}

Our models are trained on the training set of Kinetics-400~\citep{kay2017kinetics}, without using any labels, and tested on HMDB-51~\citep{kuehne2011hmdb}, UCF-101~\citep{soomro2012ucf101},  Kinetics-600~\citep{carreira2018short} and the validation set of Kinetics-400. For all the experiments, unless otherwise stated, we sample uniformly 8 frames at a resolution of $224\times 224$px. During inference we follow BLIP's approach and use beam search to generate the captions.

\textbf{Zero-shot experiments}: We train our model following Algorithm~\ref{alg:method} on the $\sim200K$ videos comprising the training set of Kinetics-400, without using any labels. We initialize the algorithm with the following parameters: the number of nearest neighbours $H$ is set to 2,000, the size of the pseudo-caption cache per video is $K=3$, and the number of epochs $R$ between retrievals is set to 10. The model is trained for 60 epochs. The algorithm and the models were implemented using PyTorch~\citep{paszke2019pytorch}. The full list of hyper-parameters is reported in the appendix. In all cases, we evaluate the accuracy of our method using our newly introduced CLIP-TAM. 

\textit{Evaluation protocol}: On HMDB-51 and UCF-101, following~\citet{ni2022expanding}, we report the average top-1 accuracy and standard deviation computed across each of the three test splits. Similarly, on Kinetics-600, we perform evaluations on the three testing splits introduced in~\citet{chen2021elaborative}. Each split contains videos belonging to 160 classes different from the ones found in Kinetics-400, with a total of 220 unique classes among the three splits. Note that~\citet{chen2021elaborative} reassigned the labels to ensure that there is no overlap between the defined 220 classes and the 400 ones from Kinetics-400. Further to this standard setting, we also report results for the more challenging generalized zero-shot setting. Finally, we also report accuracy on the validation set of Kinetics-400.

\textbf{Few-shot experiments:} We also validate our model for few-shot recognition tasks. We finetune the trained models directly using the class names (instead of action-specific captions). Depending on the number of samples per class available at train time, we finetune the model between 50 (for 16-shot) and 200 epochs (for 2-shot) using the same hyper-parameters used as for the REST training.

\textit{Evaluation protocol}: We apply the standard $M$-shot setting with $M=\{2, 4, 8, 16\}$ training videos per class. We report the average Top-1 and Top-5 accuracy on the UCF-101 and HMDB-51 datasets, after training and evaluating on each of the 3 train-test splits.

\subsection{Comparison with state-of-the-art}
\label{ssec:sota}

\textbf{Evaluation of different types of supervision:} In Table~\ref{tab:zero_shot_k400}, we compare three types of supervision on the validation set of Kinetics-400. CLIP and BLIP are trained on very large scale datasets with image-language supervision. For BLIP, we evaluate its performance in standard generative mode but also in \textit{CLIP mode}. A6~\citep{ju2022prompting} and X-CLIP~\citep{ni2022expanding} are methods based on CLIP adaptation trained in a fully supervised manner on Kinetics-400. Our model was also trained on Kinetics-400 but without any labels in an unsupervised manner. As expected A6 and X-CLIP perform the best. Notably, our method significantly outperforms both CLIP~\citep{radford2021learning} and BLIP~\citep{li2022blip}, illustrating the impact of the unsupervised adaptation on the target dataset.

Another interesting observation is that BLIP, in both generative and CLIP mode, performs significantly worse than CLIP. This shows that REST, initialized with BLIP, is in a disadvantage when compared with other methods based on CLIP adaptation (e.g.~\cite{ju2022prompting, ni2022expanding}).

\begin{table}[ht]
    \centering
        \caption{Evaluation of different types of supervision.}
    \label{tab:zero_shot_k400}
    \begin{tabular}{lccc}
    \toprule
         Method & Dataset & Top-1 & Top-5 \\
         \midrule
         \multicolumn{4}{l}{\textit{Web image-language supervision}} \\
         CLIP~\citep{radford2021learning} & CLIP-400M & \textbf{53.6} & \textbf{79.9} \\
         BLIP (CLIP mode)~\citep{li2022blip} & LAION-115M & 47.3 & 66.1\\
         BLIP~\citep{li2022blip}  &LAION-115M  & 23.5 & 41.7\\
         \midrule
         \multicolumn{4}{l}{\textit{Full supervision}} \\
         A6~\citep{ju2022prompting} & Kinetics-400 & 76.9 & 93.5 \\
         X-CLIP~\citep{ni2022expanding} & Kinetics-400  & \textbf{83.8} & \textbf{96.7}  \\
         \midrule
         \multicolumn{4}{l}{\textit{Unsupervised adaptation}} \\
         REST (Ours)  & Kinetics-400  & \textbf{63.9} & \textbf{81.0} \\
         \bottomrule
    \end{tabular}
\end{table}

\textbf{Evaluation on HMDB-51/UCF-101:} Zero-shot classification results are shown in Table~\ref{tab:zero_shot_ucf101_hmdb}, where we compare the performance of our approach against current state-of-the-art methods, including the concurrent works of~\citet{ni2022expanding,ge2022bridging}. Despite the generative nature of our approach, our method sets a new state-of-the-art result on HMDB-51, and delivers competitive results on UCF-101 where the best method is \citet{ni2022expanding} trained in a supervised setting. 

\textbf{Evaluation on Kinetics-220:} We report the results of REST against state-of-the-art methods in Tables~\ref{tab:zero_shot_k600} and~\ref{tab:zero_shot_k600_generalized} under \textit{two different scenarios}. In Table~\ref{tab:zero_shot_k600}, we report the results obtained under the standard setting of \textit{novel, unseen classes} where the evaluation is done over the restricted set of novel classes only, i.e. where the models have to perform $1\text{-vs-}160$ classification\footnote{While there are a total of $220$ unique classes in Kinetics-220 each split contains $160$ only.}. Under this setting, our model is only surpassed by the concurrent work of \citep{ni2022expanding}. 

Table~\ref{tab:zero_shot_k600_generalized} shows results for the more challenging \textit{generalized zero-shot} setting, i.e. where the model is evaluated under a $1\text{-vs-}620$ classification setting. For this setting, not surprisingly, CLIP, due to being trained on a very large dataset, performs the best. Notably, our method largely outperforms~\citet{ni2022expanding}, showcasing the benefits of generative modelling for zero-shot action recognition. 

\textbf{Few-shot experiments:} Results are reported in Table~\ref{tab:zero_shot_sota_comparison}. Notably, our approach gets a large boost with minimum additional training, surpassing the concurrent work of \cite{ni2022expanding} when $M=2$ and $M=4$, as well as getting on par results for $M=8$ and $M=16$.

\begin{table}[ht]
    \centering
        \caption{Zero-shot classification results on HMDB-51 and UCF-101 in terms of top-1 accuracy.}
    \label{tab:zero_shot_ucf101_hmdb}
    \begin{tabular}{lcc}
    \toprule
         Method & HMDB-51 & UCF-101 \\
         \midrule
         \multicolumn{3}{c}{Discriminative approaches} \\
         \midrule
         E2E~\citep{gao2019know}  & $32.7$  & $48$ \\
         TS-GCN~\citep{brattoli2020rethinking}  & $23.2 \pm 3.0$ & $34.2\pm 3.1$ \\
         ER-ZSAR~\citep{chen2021elaborative}  & $35.3 \pm 4.6$ & $51.8 \pm 2.9$ \\
         CLIP~\citep{radford2021learning}  & 46.2 & 73.0 \\
         MUFI~\citep{qiu2021boosting}  & 31.0 & 60.9 \\
         ActionCLIP~\citep{wang2021actionclip}  & $40.8\pm5.4$ & $58.3\pm 3.4$ \\
         ClipBert~\citep{lei2021less}  & $21.4\pm1.0$ & $27.8\pm 0.8$ \\
         Frozen~\citep{bain2021frozen}  & $27.8\pm0.3$ & $45.9\pm 1.3$ \\
         ViSET-96~\citep{doshi2022zero}  & $40.2$ & $68.3$ \\
         BridgeFormer~\citep{ge2022bridging}   & $37.7\pm1.2$ & $53.1\pm 1.4$\\
         AURL~\citep{pu2022alignment}  & $40.4$ & $60.9$ \\
         ResT\_101~\citep{lin2022cross}  & $41.1 \pm 3.7$ & $58.7 \pm 3.3$ \\
         X-CLIP~\citep{ni2022expanding}  & $44.6 \pm 5.2$ & $72 \pm 2.3$ \\
         X-Florence~\citep{ni2022expanding}  & $48.4 \pm 4.9$ & $\mathbf{73.2 \pm 4.2}$ \\
         
         \midrule
          \multicolumn{3}{c}{Generative approaches} \\
          \midrule
         REST (Ours)  & $\mathbf{49.7 \pm 1.14}$ & $69.1 \pm$ 0.62 \\
         \bottomrule
    \end{tabular}
\end{table}

\begin{table}[ht]
    \centering
        \caption{Zero-shot classification results on Kinetics-220 ($1\text{-vs-}160$ setting).}
    \label{tab:zero_shot_k600}
    \begin{tabular}{lcc}
    \toprule
         Method & Top-1 & Top-5 \\
         \midrule
         \multicolumn{3}{c}{Discriminative approaches} \\
         \midrule
        DEM~\citep{zhang2017learning}  & $23.6 \pm 0.6$ & $49.5\pm 0.4$ \\
        GCN~\citep{ghosh2020all}  & $22.3\pm 0.6$ & $49.7 \pm 0.6$ \\
         ER-ZSAR~\citep{chen2021elaborative} & $42.1 \pm 1.4$ & $73.1 \pm 0.3$ \\
           X-CLIP~\citep{ni2022expanding}  & $65.2 \pm 0.4$ & $86.1 \pm 0.8$ \\
         X-Florence~\citep{ni2022expanding}  & $\mathbf{68.8 \pm 0.9}$ & $\mathbf{88.4 \pm 0.6}$ \\     
         \midrule
          \multicolumn{3}{c}{Generative approaches} \\
          \midrule
         REST (Ours)  & $51.7\pm1.1$ & $75.2\pm0.4$ \\

         \bottomrule
    \end{tabular}
\end{table}

\begin{table}[!ht]
    \centering
        \caption{\textbf{Generalized} zero-shot classification results on Kinetics-220 (1-vs-620 setting).}
    \label{tab:zero_shot_k600_generalized}
    \begin{tabular}{lcc}
    \toprule
         Method   & Top-1 & Top-5 \\
         \midrule
         \multicolumn{3}{c}{Discriminative approaches} \\
         \midrule
           CLIP~\citep{radford2021learning}  & $\mathbf{47.03}$ & $\mathbf{74.4}$ \\
           X-CLIP~\citep{ni2022expanding}  & $14.76\pm0.51$ & $60.93\pm0.25$ \\
         \midrule
          \multicolumn{3}{c}{Generative approaches} \\
          \midrule
         REST (Ours)  & $29.51\pm0.71$ & $56.12\pm0.37$ \\
         \bottomrule
    \end{tabular}
    \vspace*{-0.2cm}
\end{table}

\begin{table}[!ht]
    \centering
        \caption{Few-shot classification results on HMDB-51 and UCF-101 in terms of top-1 accuracy.}
    \label{tab:zero_shot_sota_comparison}
    \begin{tabular}{l cccc| cccc}
    \toprule
         \multirow{2}{*}{Method}  & \multicolumn{4}{c}{HMDB-51} & \multicolumn{4}{c}{UCF-101} \\
         \cmidrule{2-9}
          & 2 & 4 & 8 & 16 & 2 & 4 & 8 & 16 \\
         \midrule
         \multicolumn{9}{c}{Discriminative approaches} \\
         \midrule
         TSM~\citep{lin2019tsm}  & 17.5 & 20.9 & 18.4 & 31.0 & 25.3 & 47.0 & 64.4 & 61.0 \\
         TimeSformer~\citep{bertasius2021space}  & 19.6 & 40.6 & 49.4 & 55.4 & 48.5 & 75.6 & 83.7 & 89.4 \\
         Swin-B~\citep{liu2022video}  & 20.9 & 41.3 & 47.9 & 56.1 & 53.3 & 74.1 & 85.8 & 88.7 \\
         X-CLIP~\citep{ni2022expanding}  & 53.0 & 57.3 & 62.8 & 64.0 & 76.4 & 83.4 & 88.3 & 91.4 \\
         X-Florence~\citep{ni2022expanding}  & 51.6 & 57.8 & \textbf{64.1} & \textbf{64.2} & 84.0 & 88.5 & 92.5 & \textbf{94.8} \\
         \midrule
          \multicolumn{9}{c}{Generative approaches} \\
          \midrule
         REST (Ours) &  \textbf{54.0} & \textbf{59.1} &  62.1 & 64.0 &  \textbf{88.2} & \textbf{90.2} &  \textbf{92.6} & 93.5\\
         \bottomrule
         \vspace*{-0.5cm}
    \end{tabular}

\end{table}

\subsection{Ablation studies}

\textbf{Effect of retrieval step:} A key component of REST is integrating retrieval into self-training. We analyze its importance by conducting an experiment where only self-training is used (i.e. $H=1$). We observe that the results on the validation set of Kinetics-400 drop massively from $63.9\%$ to $44.0\%$, clearly validating the importance of retrieval in REST. 

\textbf{Effect of number of cached captions:} We evaluate the impact of $K$ in Algorithm~\ref{alg:method} (the number of generated captions cached per video) on Kinetics-400. The results shown in Table~\ref{tab:zero_shot_k400_K} suggest that increasing $K$ from 1 to 3 offers a significant improvement in terms of Top-1 accuracy. However, going beyond $K=3$ does not offer further gains indicating that samples ranked lower are less likely to be representative of the video.

\begin{table}[ht]
    \centering
    \caption{Analysis on Kinetics-400.}
    \label{tab:analysis}
\begin{subtable}[t]{.4\textwidth}
    \centering
        \caption{Effect of number of cached captions.}
    \label{tab:zero_shot_k400_K}
    \begin{tabular}{lccc}
    \toprule
          & K = 1 & K = 3 & K = 5 \\
         \midrule
         Ours  & 62.60 & \textbf{63.93} & \textbf{63.98} \\
         \bottomrule
    \end{tabular}
\end{subtable}
\hfill
\begin{subtable}[t]{.5\textwidth}
    \centering
        \caption{Effect of number of retrieval steps.}
    \label{tab:zero_shot_k400_retrievals}
    \begin{tabular}{lcccc}
    \toprule
          & $N_I$ = 1 & $N_I$ = 2 & $N_I$ = 3 & $N_I$ = 4 \\
         \midrule
         Ours  & 59.16 & 62.75 & \textbf{63.93} & \textbf{63.95}\\
         \bottomrule
    \end{tabular}
\end{subtable}
\end{table}

\textbf{Effect of number of retrieval steps:} We evaluate the effect of the frequency of updating the pseudo-captions as shown in l.10-13 of Algorithm~\ref{alg:method}.  Table~\ref{tab:zero_shot_k400_retrievals} shows that updating the captions results in better performance, showing that, over the course of training, they become more semantically meaningful. Generally, up to a certain point, more retrieval steps correlate with better performance.

\section{Conclusion}

In this paper, we proposed REST, the very first generative method for video/action recognition trained without human supervision. Departing from an image-based generative model, an iterative algorithm that alternates between self-training and retrieving is proposed, leading to a model that can produce captions with a strong semantic correlation with action classification. Remarkably, our model can operate in a zero-shot setting without the need of manually defining a target set, and sets a strong baseline for adaptation with limited data. We believe our method sets a new path on training generative models for zero-shot action recognition in a truly open-set scenario. Code and models will be released to encourage further analysis.

\section*{Ethics considerations}

Generative language models exhibit various forms of biases learned from the data, such as occupational biases that link certain activities with some groups of individuals~\citep{kirk2021bias}. Our work derives it pseudo-labels from the representation learned by BLIP and CLIP models, and hence, the resulting network may have inherited the bias present in the source. Therefore, before deployment, the models should undergo in-depth checks and considerations.

\section*{Reproducibility statement}

We detail in Section~\ref{ssec:experimental setting} the training settings and evaluation protocols, while in appendix, Table~\ref{tab:hyper-parameters} we list the augmentations and training hyper-parameters. We will also release the code to ensure the reproducibility of our approach.

\bibliography{iclr2023_conference}

\begin{thebibliography}{64}
\providecommand{\natexlab}[1]{#1}
\providecommand{\url}[1]{\texttt{#1}}
\expandafter\ifx\csname urlstyle\endcsname\relax
  \providecommand{\doi}[1]{doi: #1}\else
  \providecommand{\doi}{doi: \begingroup \urlstyle{rm}\Url}\fi

\bibitem[Alayrac et~al.(2022)Alayrac, Donahue, Luc, Miech, Barr, Hasson, Lenc,
  Mensch, Millican, Reynolds, et~al.]{alayrac2022flamingo}
Jean-Baptiste Alayrac, Jeff Donahue, Pauline Luc, Antoine Miech, Iain Barr,
  Yana Hasson, Karel Lenc, Arthur Mensch, Katie Millican, Malcolm Reynolds,
  et~al.
\newblock Flamingo: a visual language model for few-shot learning.
\newblock \emph{arXiv preprint arXiv:2204.14198}, 2022.

\bibitem[Arazo et~al.(2020)Arazo, Ortego, Albert, O’Connor, and
  McGuinness]{arazo2020pseudo}
Eric Arazo, Diego Ortego, Paul Albert, Noel~E O’Connor, and Kevin McGuinness.
\newblock Pseudo-labeling and confirmation bias in deep semi-supervised
  learning.
\newblock In \emph{International Joint Conference on Neural Networks}, 2020.

\bibitem[Bain et~al.(2021)Bain, Nagrani, Varol, and Zisserman]{bain2021frozen}
Max Bain, Arsha Nagrani, G{\"u}l Varol, and Andrew Zisserman.
\newblock Frozen in time: A joint video and image encoder for end-to-end
  retrieval.
\newblock In \emph{IEEE International Conference on Computer Vision}, 2021.

\bibitem[Bertasius et~al.(2021)Bertasius, Wang, and
  Torresani]{bertasius2021space}
Gedas Bertasius, Heng Wang, and Lorenzo Torresani.
\newblock Is space-time attention all you need for video understanding?
\newblock In \emph{International Conference on Machine Learning}, 2021.

\bibitem[Brattoli et~al.(2020)Brattoli, Tighe, Zhdanov, Perona, and
  Chalupka]{brattoli2020rethinking}
Biagio Brattoli, Joseph Tighe, Fedor Zhdanov, Pietro Perona, and Krzysztof
  Chalupka.
\newblock Rethinking zero-shot video classification: End-to-end training for
  realistic applications.
\newblock In \emph{IEEE Conference on Computer Vision and Pattern Recognition},
  2020.

\bibitem[Bretti \& Mettes(2021)Bretti and Mettes]{bretti2021zero}
Carlo Bretti and Pascal Mettes.
\newblock Zero-shot action recognition from diverse object-scene compositions.
\newblock \emph{arXiv preprint arXiv:2110.13479}, 2021.

\bibitem[Carreira et~al.(2018)Carreira, Noland, Banki-Horvath, Hillier, and
  Zisserman]{carreira2018short}
Joao Carreira, Eric Noland, Andras Banki-Horvath, Chloe Hillier, and Andrew
  Zisserman.
\newblock A short note about kinetics-600.
\newblock \emph{arXiv preprint arXiv:1808.01340}, 2018.

\bibitem[Castro \& Heilbron(2022)Castro and Heilbron]{castro2022fitclip}
Santiago Castro and Fabian~Caba Heilbron.
\newblock Fitclip: Refining large-scale pretrained image-text models for
  zero-shot video understanding tasks.
\newblock \emph{arXiv preprint arXiv:2203.13371}, 2022.

\bibitem[Chen \& Huang(2021)Chen and Huang]{chen2021elaborative}
Shizhe Chen and Dong Huang.
\newblock Elaborative rehearsal for zero-shot action recognition.
\newblock In \emph{IEEE International Conference on Computer Vision}, 2021.

\bibitem[Chen et~al.(2017)Chen, Liao, Chuang, Hsu, Fu, and Sun]{chen2017show}
Tseng-Hung Chen, Yuan-Hong Liao, Ching-Yao Chuang, Wan-Ting Hsu, Jianlong Fu,
  and Min Sun.
\newblock Show, adapt and tell: Adversarial training of cross-domain image
  captioner.
\newblock In \emph{IEEE International Conference on Computer Vision}, 2017.

\bibitem[Chen et~al.(2021)Chen, Jiang, and Zhao]{chen2021self}
Xianyu Chen, Ming Jiang, and Qi~Zhao.
\newblock Self-distillation for few-shot image captioning.
\newblock In \emph{Winter Conference on Applications of Computer Vision}, 2021.

\bibitem[Doshi \& Yilmaz(2022)Doshi and Yilmaz]{doshi2022zero}
Keval Doshi and Yasin Yilmaz.
\newblock Zero-shot action recognition with transformer-based video semantic
  embedding.
\newblock \emph{arXiv preprint arXiv:2203.05156}, 2022.

\bibitem[Dosovitskiy et~al.(2020)Dosovitskiy, Beyer, Kolesnikov, Weissenborn,
  Zhai, Unterthiner, Dehghani, Minderer, Heigold, Gelly,
  et~al.]{dosovitskiy2020image}
Alexey Dosovitskiy, Lucas Beyer, Alexander Kolesnikov, Dirk Weissenborn,
  Xiaohua Zhai, Thomas Unterthiner, Mostafa Dehghani, Matthias Minderer, Georg
  Heigold, Sylvain Gelly, et~al.
\newblock An image is worth 16x16 words: Transformers for image recognition at
  scale.
\newblock \emph{arXiv preprint arXiv:2010.11929}, 2020.

\bibitem[Du et~al.(2022)Du, Wei, Zhang, Shi, Gao, and Li]{du2022learning}
Yu~Du, Fangyun Wei, Zihe Zhang, Miaojing Shi, Yue Gao, and Guoqi Li.
\newblock Learning to prompt for open-vocabulary object detection with
  vision-language model.
\newblock In \emph{IEEE Conference on Computer Vision and Pattern Recognition},
  2022.

\bibitem[Estevam et~al.(2022)Estevam, Laroca, Pedrini, and
  Menotti]{estevam2022global}
Valter Estevam, Rayson Laroca, Helio Pedrini, and David Menotti.
\newblock Global semantic descriptors for zero-shot action recognition.
\newblock \emph{IEEE Signal Processing Letters}, 2022.

\bibitem[Feng et~al.(2019)Feng, Ma, Liu, and Luo]{feng2019unsupervised}
Yang Feng, Lin Ma, Wei Liu, and Jiebo Luo.
\newblock Unsupervised image captioning.
\newblock In \emph{IEEE Conference on Computer Vision and Pattern Recognition},
  2019.

\bibitem[Gao et~al.(2019)Gao, Zhang, and Xu]{gao2019know}
Junyu Gao, Tianzhu Zhang, and Changsheng Xu.
\newblock I know the relationships: Zero-shot action recognition via two-stream
  graph convolutional networks and knowledge graphs.
\newblock In \emph{AAAI Conference on Artificial Intelligence}, 2019.

\bibitem[Ge et~al.(2022)Ge, Ge, Liu, Li, Shan, Qie, and Luo]{ge2022bridging}
Yuying Ge, Yixiao Ge, Xihui Liu, Dian Li, Ying Shan, Xiaohu Qie, and Ping Luo.
\newblock Bridging video-text retrieval with multiple choice questions.
\newblock In \emph{IEEE Conference on Computer Vision and Pattern Recognition},
  2022.

\bibitem[Ghosh et~al.(2020)Ghosh, Saini, Davis, and Shrivastava]{ghosh2020all}
Pallabi Ghosh, Nirat Saini, Larry~S Davis, and Abhinav Shrivastava.
\newblock All about knowledge graphs for actions.
\newblock \emph{arXiv preprint arXiv:2008.12432}, 2020.

\bibitem[Gu et~al.(2021)Gu, Lin, Kuo, and Cui]{gu2021open}
Xiuye Gu, Tsung-Yi Lin, Weicheng Kuo, and Yin Cui.
\newblock Open-vocabulary object detection via vision and language knowledge
  distillation.
\newblock \emph{arXiv preprint arXiv:2104.13921}, 2021.

\bibitem[Jia et~al.(2021)Jia, Yang, Xia, Chen, Parekh, Pham, Le, Sung, Li, and
  Duerig]{jia2021scaling}
Chao Jia, Yinfei Yang, Ye~Xia, Yi-Ting Chen, Zarana Parekh, Hieu Pham, Quoc Le,
  Yun-Hsuan Sung, Zhen Li, and Tom Duerig.
\newblock Scaling up visual and vision-language representation learning with
  noisy text supervision.
\newblock In \emph{International Conference on Machine Learning}, 2021.

\bibitem[Ju et~al.(2022)Ju, Han, Zheng, Zhang, and Xie]{ju2022prompting}
Chen Ju, Tengda Han, Kunhao Zheng, Ya~Zhang, and Weidi Xie.
\newblock Prompting visual-language models for efficient video understanding.
\newblock \emph{European Conference on Computer Vision}, 2022.

\bibitem[Kay et~al.(2017)Kay, Carreira, Simonyan, Zhang, Hillier,
  Vijayanarasimhan, Viola, Green, Back, Natsev, et~al.]{kay2017kinetics}
Will Kay, Joao Carreira, Karen Simonyan, Brian Zhang, Chloe Hillier, Sudheendra
  Vijayanarasimhan, Fabio Viola, Tim Green, Trevor Back, Paul Natsev, et~al.
\newblock The {Kinetics} human action video dataset.
\newblock \emph{arXiv preprint arXiv:1705.06950}, 2017.

\bibitem[Kim et~al.(2019)Kim, Choi, Oh, and Kweon]{kim2019image}
Dong-Jin Kim, Jinsoo Choi, Tae-Hyun Oh, and In~So Kweon.
\newblock Image captioning with very scarce supervised data: Adversarial
  semi-supervised learning approach.
\newblock \emph{arXiv preprint arXiv:1909.02201}, 2019.

\bibitem[Kirk et~al.(2021)Kirk, Jun, Volpin, Iqbal, Benussi, Dreyer,
  Shtedritski, and Asano]{kirk2021bias}
Hannah~Rose Kirk, Yennie Jun, Filippo Volpin, Haider Iqbal, Elias Benussi,
  Frederic Dreyer, Aleksandar Shtedritski, and Yuki Asano.
\newblock Bias out-of-the-box: An empirical analysis of intersectional
  occupational biases in popular generative language models.
\newblock \emph{Advances on Neural Information Processing Systems}, 2021.

\bibitem[Kuehne et~al.(2011)Kuehne, Jhuang, Garrote, Poggio, and
  Serre]{kuehne2011hmdb}
Hildegard Kuehne, Hueihan Jhuang, Est{\'\i}baliz Garrote, Tomaso Poggio, and
  Thomas Serre.
\newblock {HMDB}: a large video database for human motion recognition.
\newblock In \emph{IEEE International Conference on Computer Vision}, 2011.

\bibitem[Laina et~al.(2019)Laina, Rupprecht, and Navab]{laina2019towards}
Iro Laina, Christian Rupprecht, and Nassir Navab.
\newblock Towards unsupervised image captioning with shared multimodal
  embeddings.
\newblock In \emph{IEEE International Conference on Computer Vision}, 2019.

\bibitem[Lei et~al.(2021)Lei, Li, Zhou, Gan, Berg, Bansal, and
  Liu]{lei2021less}
Jie Lei, Linjie Li, Luowei Zhou, Zhe Gan, Tamara~L Berg, Mohit Bansal, and
  Jingjing Liu.
\newblock Less is more: Clipbert for video-and-language learning via sparse
  sampling.
\newblock In \emph{IEEE Conference on Computer Vision and Pattern Recognition},
  2021.

\bibitem[Levinboim et~al.(2019)Levinboim, Thapliyal, Sharma, and
  Soricut]{levinboim2019quality}
Tomer Levinboim, Ashish~V Thapliyal, Piyush Sharma, and Radu Soricut.
\newblock Quality estimation for image captions based on large-scale human
  evaluations.
\newblock \emph{arXiv preprint arXiv:1909.03396}, 2019.

\bibitem[Li et~al.(2022)Li, Li, Xiong, and Hoi]{li2022blip}
Junnan Li, Dongxu Li, Caiming Xiong, and Steven Hoi.
\newblock Blip: Bootstrapping language-image pre-training for unified
  vision-language understanding and generation.
\newblock \emph{arXiv preprint arXiv:2201.12086}, 2022.

\bibitem[Li et~al.(2021)Li, Liang, Zhao, Cui, Ouyang, Shao, Yu, and
  Yan]{li2021supervision}
Yangguang Li, Feng Liang, Lichen Zhao, Yufeng Cui, Wanli Ouyang, Jing Shao,
  Fengwei Yu, and Junjie Yan.
\newblock Supervision exists everywhere: A data efficient contrastive
  language-image pre-training paradigm.
\newblock \emph{arXiv preprint arXiv:2110.05208}, 2021.

\bibitem[Lin(2004)]{lin2004rouge}
Chin-Yew Lin.
\newblock Rouge: A package for automatic evaluation of summaries.
\newblock In \emph{Text summarization branches out}, pp.\  74--81, 2004.

\bibitem[Lin et~al.(2022{\natexlab{a}})Lin, Lin, Wang, Liu, and
  Li]{lin2022cross}
Chung-Ching Lin, Kevin Lin, Lijuan Wang, Zicheng Liu, and Linjie Li.
\newblock Cross-modal representation learning for zero-shot action recognition.
\newblock In \emph{IEEE Conference on Computer Vision and Pattern Recognition},
  2022{\natexlab{a}}.

\bibitem[Lin et~al.(2019)Lin, Gan, and Han]{lin2019tsm}
Ji~Lin, Chuang Gan, and Song Han.
\newblock {TSM}: Temporal shift module for efficient video understanding.
\newblock In \emph{IEEE International Conference on Computer Vision}, 2019.

\bibitem[Lin et~al.(2022{\natexlab{b}})Lin, Li, Lin, Ahmed, Gan, Liu, Lu, and
  Wang]{lin2021end-to-end}
Kevin Lin, Linjie Li, Chung-Ching Lin, Faisal Ahmed, Zhe Gan, Zicheng Liu,
  Yumao Lu, and Lijuan Wang.
\newblock {SwinBERT}: End-to-end transformers with sparse attention for video
  captioning.
\newblock In \emph{IEEE Conference on Computer Vision and Pattern Recognition},
  2022{\natexlab{b}}.

\bibitem[Lin et~al.(2022{\natexlab{c}})Lin, Geng, Zhang, Gao, de~Melo, Wang,
  Dai, Qiao, and Li]{lin2022frozen}
Ziyi Lin, Shijie Geng, Renrui Zhang, Peng Gao, Gerard de~Melo, Xiaogang Wang,
  Jifeng Dai, Yu~Qiao, and Hongsheng Li.
\newblock Frozen {CLIP} models are efficient video learners.
\newblock \emph{arXiv preprint arXiv:2208.03550}, 2022{\natexlab{c}}.

\bibitem[Liu et~al.(2022)Liu, Ning, Cao, Wei, Zhang, Lin, and Hu]{liu2022video}
Ze~Liu, Jia Ning, Yue Cao, Yixuan Wei, Zheng Zhang, Stephen Lin, and Han Hu.
\newblock Video swin transformer.
\newblock In \emph{IEEE Conference on Computer Vision and Pattern Recognition},
  2022.

\bibitem[Loshchilov \& Hutter(2017)Loshchilov and
  Hutter]{loshchilov2017decoupled}
Ilya Loshchilov and Frank Hutter.
\newblock Decoupled weight decay regularization.
\newblock \emph{arXiv preprint arXiv:1711.05101}, 2017.

\bibitem[Luo et~al.(2022)Luo, Ghosh, Guillory, Kato, Darrell, and
  Xu]{luo2022disentangled}
Zhekun Luo, Shalini Ghosh, Devin Guillory, Keizo Kato, Trevor Darrell, and
  Huijuan Xu.
\newblock Disentangled action recognition with knowledge bases.
\newblock In \emph{Conference of the North American Chapter of the Association
  for Computational Linguistics: Human Language Technologies}, 2022.

\bibitem[Mettes(2022)]{mettes2022universal}
Pascal Mettes.
\newblock Universal prototype transport for zero-shot action recognition and
  localization.
\newblock \emph{arXiv preprint arXiv:2203.03971}, 2022.

\bibitem[Minderer et~al.(2022)Minderer, Gritsenko, Stone, Neumann, Weissenborn,
  Dosovitskiy, Mahendran, Arnab, Dehghani, Shen, et~al.]{minderer2022simple}
Matthias Minderer, Alexey Gritsenko, Austin Stone, Maxim Neumann, Dirk
  Weissenborn, Alexey Dosovitskiy, Aravindh Mahendran, Anurag Arnab, Mostafa
  Dehghani, Zhuoran Shen, et~al.
\newblock Simple open-vocabulary object detection with vision transformers.
\newblock \emph{arXiv preprint arXiv:2205.06230}, 2022.

\bibitem[Ni et~al.(2022)Ni, Peng, Chen, Zhang, Meng, Fu, Xiang, and
  Ling]{ni2022expanding}
Bolin Ni, Houwen Peng, Minghao Chen, Songyang Zhang, Gaofeng Meng, Jianlong Fu,
  Shiming Xiang, and Haibin Ling.
\newblock Expanding language-image pretrained models for general video
  recognition.
\newblock \emph{arXiv preprint arXiv:2208.02816}, 2022.

\bibitem[Pan et~al.(2020)Pan, Cai, Huang, Lee, Gaidon, Adeli, and
  Niebles]{pan2020spatio}
Boxiao Pan, Haoye Cai, De-An Huang, Kuan-Hui Lee, Adrien Gaidon, Ehsan Adeli,
  and Juan~Carlos Niebles.
\newblock Spatio-temporal graph for video captioning with knowledge
  distillation.
\newblock In \emph{IEEE Conference on Computer Vision and Pattern Recognition},
  2020.

\bibitem[Papineni et~al.(2002)Papineni, Roukos, Ward, and
  Zhu]{papineni2002bleu}
Kishore Papineni, Salim Roukos, Todd Ward, and Wei-Jing Zhu.
\newblock Bleu: a method for automatic evaluation of machine translation.
\newblock In \emph{Annual meeting of the Association for Computational
  Linguistics}, 2002.

\bibitem[Paszke et~al.(2019)Paszke, Gross, Massa, Lerer, Bradbury, Chanan,
  Killeen, Lin, Gimelshein, Antiga, et~al.]{paszke2019pytorch}
Adam Paszke, Sam Gross, Francisco Massa, Adam Lerer, James Bradbury, Gregory
  Chanan, Trevor Killeen, Zeming Lin, Natalia Gimelshein, Luca Antiga, et~al.
\newblock Pytorch: An imperative style, high-performance deep learning library.
\newblock \emph{Advances on Neural Information Processing Systems}, 2019.

\bibitem[Pu et~al.(2022)Pu, Zhao, and Zheng]{pu2022alignment}
Shi Pu, Kaili Zhao, and Mao Zheng.
\newblock Alignment-uniformity aware representation learning for zero-shot
  video classification.
\newblock In \emph{IEEE Conference on Computer Vision and Pattern Recognition},
  2022.

\bibitem[Qian et~al.(2022)Qian, Li, Xu, Yang, Belongie, and
  Cui]{qian2022multimodal}
Rui Qian, Yeqing Li, Zheng Xu, Ming-Hsuan Yang, Serge Belongie, and Yin Cui.
\newblock Multimodal open-vocabulary video classification via pre-trained
  vision and language models.
\newblock \emph{arXiv preprint arXiv:2207.07646}, 2022.

\bibitem[Qiu et~al.(2021)Qiu, Yao, Ngo, Zhang, Wu, and Mei]{qiu2021boosting}
Zhaofan Qiu, Ting Yao, Chong-Wah Ngo, Xiao-Ping Zhang, Dong Wu, and Tao Mei.
\newblock Boosting video representation learning with multi-faceted
  integration.
\newblock In \emph{IEEE Conference on Computer Vision and Pattern Recognition},
  2021.

\bibitem[Radford et~al.(2021)Radford, Kim, Hallacy, Ramesh, Goh, Agarwal,
  Sastry, Askell, Mishkin, Clark, et~al.]{radford2021learning}
Alec Radford, Jong~Wook Kim, Chris Hallacy, Aditya Ramesh, Gabriel Goh,
  Sandhini Agarwal, Girish Sastry, Amanda Askell, Pamela Mishkin, Jack Clark,
  et~al.
\newblock Learning transferable visual models from natural language
  supervision.
\newblock In \emph{International Conference on Machine Learning}, 2021.

\bibitem[Soomro et~al.(2012)Soomro, Zamir, and Shah]{soomro2012ucf101}
Khurram Soomro, Amir~Roshan Zamir, and Mubarak Shah.
\newblock {UCF101}: A dataset of 101 human actions classes from videos in the
  wild.
\newblock \emph{arXiv preprint arXiv:1212.0402}, 2012.

\bibitem[Tang et~al.(2021{\natexlab{a}})Tang, Wang, LIU, Rao, Li, and
  Li]{tang2021clip4caption}
Mingkang Tang, Zhanyu Wang, Zhenhua LIU, Fengyun Rao, Dian Li, and Xiu Li.
\newblock {CLIP4Caption}: Clip for video caption.
\newblock In \emph{ACM International Conference on Multimedia},
  2021{\natexlab{a}}.

\bibitem[Tang et~al.(2021{\natexlab{b}})Tang, Wang, Liu, Rao, Li, and
  Li]{tang2021clipplus}
Mingkang Tang, Zhanyu Wang, Zhenhua Liu, Fengyun Rao, Dian Li, and Xiu Li.
\newblock {CLIP4Caption} ++: Multi-clip for video caption.
\newblock \emph{arXiv preprint arXiv:2208.02816}, 2021{\natexlab{b}}.

\bibitem[Tsimpoukelli et~al.(2021)Tsimpoukelli, Menick, Cabi, Eslami, Vinyals,
  and Hill]{tsimpoukelli2021multimodal}
Maria Tsimpoukelli, Jacob~L Menick, Serkan Cabi, SM~Eslami, Oriol Vinyals, and
  Felix Hill.
\newblock Multimodal few-shot learning with frozen language models.
\newblock \emph{Advances on Neural Information Processing Systems}, 2021.

\bibitem[Vedantam et~al.(2015)Vedantam, Lawrence~Zitnick, and
  Parikh]{vedantam2015cider}
Ramakrishna Vedantam, C~Lawrence~Zitnick, and Devi Parikh.
\newblock Cider: Consensus-based image description evaluation.
\newblock In \emph{IEEE Conference on Computer Vision and Pattern Recognition},
  2015.

\bibitem[Wang et~al.(2022)Wang, Yang, Hu, Li, Lin, Gan, Liu, Liu, and
  Wang]{wang2022git}
Jianfeng Wang, Zhengyuan Yang, Xiaowei Hu, Linjie Li, Kevin Lin, Zhe Gan,
  Zicheng Liu, Ce~Liu, and Lijuan Wang.
\newblock {GIT}: A generative image-to-text transformer for vision and
  language.
\newblock \emph{arXiv preprint arXiv:2205.14100}, 2022.

\bibitem[Wang et~al.(2021)Wang, Xing, and Liu]{wang2021actionclip}
Mengmeng Wang, Jiazheng Xing, and Yong Liu.
\newblock Actionclip: A new paradigm for video action recognition.
\newblock \emph{arXiv preprint arXiv:2109.08472}, 2021.

\bibitem[Yan et~al.(2022)Yan, Xiong, Arnab, Lu, Zhang, Sun, and
  Schmid]{yan2022multiview}
Shen Yan, Xuehan Xiong, Anurag Arnab, Zhichao Lu, Mi~Zhang, Chen Sun, and
  Cordelia Schmid.
\newblock Multiview transformers for video recognition.
\newblock In \emph{IEEE Conference on Computer Vision and Pattern Recognition},
  2022.

\bibitem[Yao et~al.(2021)Yao, Huang, Hou, Lu, Niu, Xu, Liang, Li, Jiang, and
  Xu]{yao2021filip}
Lewei Yao, Runhui Huang, Lu~Hou, Guansong Lu, Minzhe Niu, Hang Xu, Xiaodan
  Liang, Zhenguo Li, Xin Jiang, and Chunjing Xu.
\newblock Filip: Fine-grained interactive language-image pre-training.
\newblock \emph{arXiv preprint arXiv:2111.07783}, 2021.

\bibitem[Yu et~al.(2022)Yu, Wang, Vasudevan, Yeung, Seyedhosseini, and
  Wu]{yu2022coca}
Jiahui Yu, Zirui Wang, Vijay Vasudevan, Legg Yeung, Mojtaba Seyedhosseini, and
  Yonghui Wu.
\newblock Coca: Contrastive captioners are image-text foundation models.
\newblock \emph{arXiv preprint arXiv:2205.01917}, 2022.

\bibitem[Yuan et~al.(2021)Yuan, Chen, Chen, Codella, Dai, Gao, Hu, Huang, Li,
  Li, et~al.]{yuan2021florence}
Lu~Yuan, Dongdong Chen, Yi-Ling Chen, Noel Codella, Xiyang Dai, Jianfeng Gao,
  Houdong Hu, Xuedong Huang, Boxin Li, Chunyuan Li, et~al.
\newblock Florence: A new foundation model for computer vision.
\newblock \emph{arXiv preprint arXiv:2111.11432}, 2021.

\bibitem[Zhai et~al.(2022)Zhai, Wang, Mustafa, Steiner, Keysers, Kolesnikov,
  and Beyer]{zhai2022lit}
Xiaohua Zhai, Xiao Wang, Basil Mustafa, Andreas Steiner, Daniel Keysers,
  Alexander Kolesnikov, and Lucas Beyer.
\newblock Lit: Zero-shot transfer with locked-image text tuning.
\newblock In \emph{IEEE Conference on Computer Vision and Pattern Recognition},
  2022.

\bibitem[Zhang et~al.(2017)Zhang, Xiang, and Gong]{zhang2017learning}
Li~Zhang, Tao Xiang, and Shaogang Gong.
\newblock Learning a deep embedding model for zero-shot learning.
\newblock In \emph{IEEE Conference on Computer Vision and Pattern Recognition},
  2017.

\bibitem[Zhou et~al.(2021)Zhou, Tao, and Zhang]{zhou2021triple}
Yucheng Zhou, Wei Tao, and Wenqiang Zhang.
\newblock Triple sequence generative adversarial nets for unsupervised image
  captioning.
\newblock In \emph{International Conference on Acoustics, Speech, and Signal
  Processing}, 2021.

\bibitem[Zhu et~al.(2018)Zhu, Long, Guan, Newsam, and Shao]{zhu2018towards}
Yi~Zhu, Yang Long, Yu~Guan, Shawn Newsam, and Ling Shao.
\newblock Towards universal representation for unseen action recognition.
\newblock In \emph{IEEE Conference on Computer Vision and Pattern Recognition},
  2018.

\end{thebibliography}
\bibliographystyle{iclr2023_conference}

\clearpage

\appendix

\section{Additional ablation studies}

\textbf{Fully supervised training with action class names:} While the main focus of our work is learning generative zero-shot and few-shot classifiers, nevertheless, herein we train and evaluate the generative action recognition (GAR) model trained with the original class names on Kinetics-400. The training is performed using the same hyper-parameters as the ones used for REST (see Table~\ref{tab:hyper-parameters}). As the results from Table~\ref{tab:supervised_k400} show, while competitive, GAR lags behind the current CLIP-based discriminative methods. Based on our prior results, we hypothesise that this is due to: (a) the BLIP model used in our method being weaker than CLIP, and (b) certain augmentations performed for classification being harder to apply for a generative model (e.g.: mixup, cutout etc.).

\begin{table}[ht]
    \centering
        \caption{Fully supervised classification results on Kinetics-400.}
    \label{tab:supervised_k400}
    \begin{tabular}{lccc}
    \toprule
         Method & Top-1 & Top-5 & Throughput \\
         \midrule
          \multicolumn{4}{c}{Discriminative approaches} \\
         \midrule
         Action-CLIP~\citep{wang2021actionclip} & 83.8 & 96.2 & - \\
         A6~\citep{ju2022prompting} & 76.9 & 93.5 & -\\
         MVT-H~\citep{yan2022multiview} & \textbf{89.1} & \textbf{98.2} & -\\
         X-Florence~\citep{ni2022expanding} & 86.2 & 96.6 & 6 \\
         X-CLIP~\citep{ni2022expanding} & 83.8 & 96.7 & \textbf{33} \\
            \midrule
          \multicolumn{4}{c}{Generative approaches} \\
          \midrule
         GAR  & 79.0 & 87.8 & 11 \\
         \bottomrule
    \end{tabular}
\end{table}

\textbf{Effect of temporal adapter:} In Table~\ref{tab:zero_shot_k400_ablation}, we evaluate the effect of the video adapter introduced in Sec.~\ref{ssec:GAR} by training a model that applies only temporal average pooling in the last layer.

\begin{table}[ht]
    \centering
        \caption{Effect of temporal adapter.}
    \label{tab:zero_shot_k400_ablation}
    \begin{tabular}{lcc}
    \toprule
         Method & Top-1 & Top-5 \\
         \midrule
      
         Ours (w/o) adapter  & 61.5 & 79.5 \\
         Ours  & \textbf{63.93} & \textbf{81.04} \\
         \bottomrule
    \end{tabular}
\end{table}

\textbf{Training hyper-parameters} are listed in Table~\ref{tab:hyper-parameters}.
\begin{table}[ht]
    \centering
        \caption{Training hyper-parameters. K is the number of samples per class for few-shot learning.}
    \label{tab:hyper-parameters}
    \begin{tabular}{lcc}
    \toprule
         & REST training & Few-shot finetunning \\
         \midrule
           \multicolumn{3}{c}{Optimization} \\
        \midrule
        Optimizer & \multicolumn{2}{c}{AdamW~\citep{loshchilov2017decoupled}} \\
        Optimizer betas & \multicolumn{2}{c}{(0.9, 0.98)} \\
        Batch size & \multicolumn{2}{c}{64} \\
        Weight decay & \multicolumn{2}{c}{0.001} \\
        Learning rate scheduler & \multicolumn{2}{c}{cosine} \\
        Initial learning rate & \multicolumn{2}{c}{$2e-5$} \\
        Minimal learning rate & \multicolumn{2}{c}{$2e-8$} \\
        Epochs & 60 & max(400/K, 30) \\
        \midrule
           \multicolumn{3}{c}{Data augmentation} \\
        \midrule
        Random Flip & \multicolumn{2}{c}{0.5}\\
        Multi Scale Crop & \multicolumn{2}{c}{(0.66, 0.75, 0.875, 1.0)}\\
        Color Jitter & \multicolumn{2}{c}{0.8}\\
        Gray Scale & \multicolumn{2}{c}{0.2}\\
        Label smoothing & \multicolumn{2}{c}{0.2}\\
         \bottomrule
    \end{tabular}
\end{table}

\section{Qualitative results}

\begin{figure}[ht]
    \centering
    \begin{subfigure}{\textwidth}
    \includegraphics[width=\textwidth]{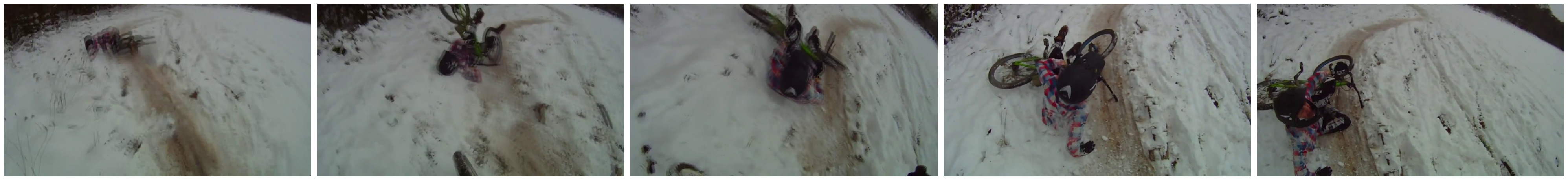}
    \vspace*{-5mm}
   \caption*{a mountain biker falling off his bike in the snow}
    \end{subfigure}    
    
    \begin{subfigure}{\textwidth}
    \includegraphics[width=\textwidth]{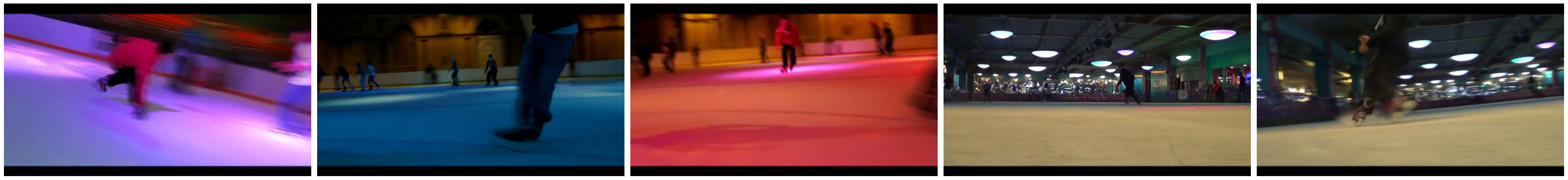}
    \vspace*{-5mm}
   \caption*{a person ice skating on an ice rink}
    \end{subfigure}    

    \begin{subfigure}{\textwidth}
    \includegraphics[width=\textwidth]{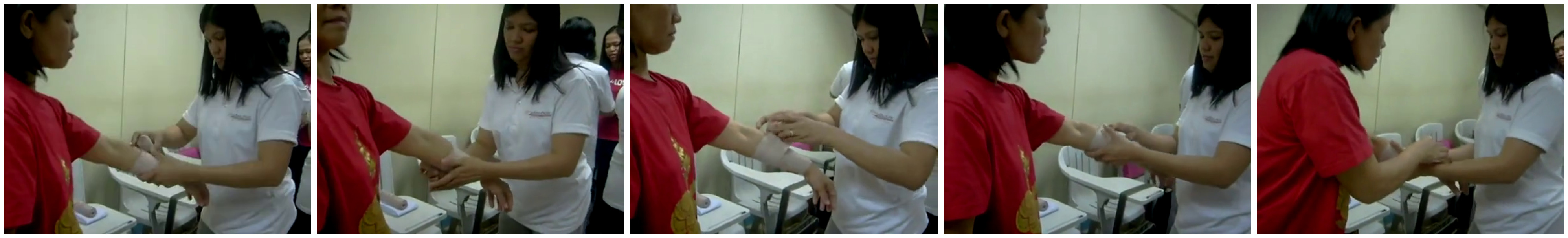}
    \vspace*{-5mm}
   \caption*{a nurse putting a bandage on a patient's arm}
    \end{subfigure}    

    \begin{subfigure}{\textwidth}
    \includegraphics[width=\textwidth]{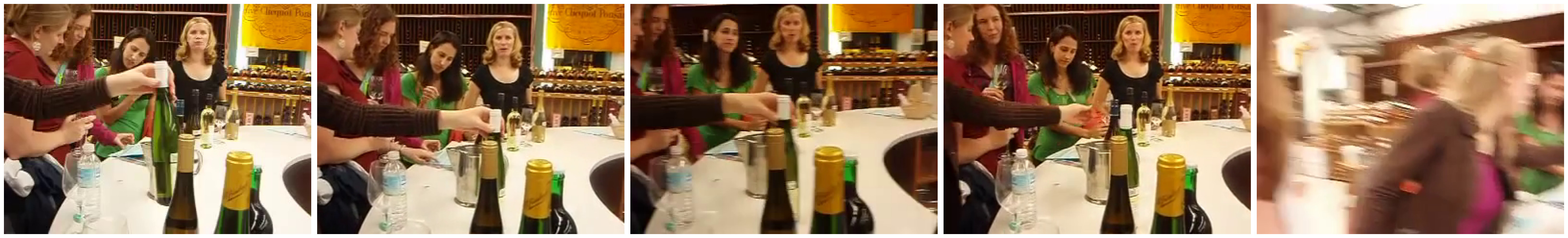}
    \vspace*{-5mm}
   \caption*{a group of friends drinking wine and talking to each other while sitting at a table in a restaurant}
    \end{subfigure}    

    \begin{subfigure}{\textwidth}
    \includegraphics[width=\textwidth]{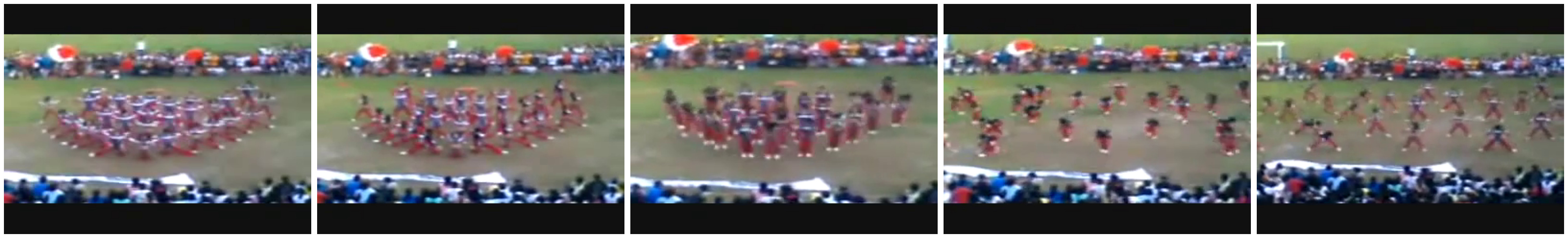}
    \vspace*{-5mm}
   \caption*{a cheer team performing on a football field}
    \end{subfigure} 

    \begin{subfigure}{\textwidth}
    \includegraphics[width=\textwidth]{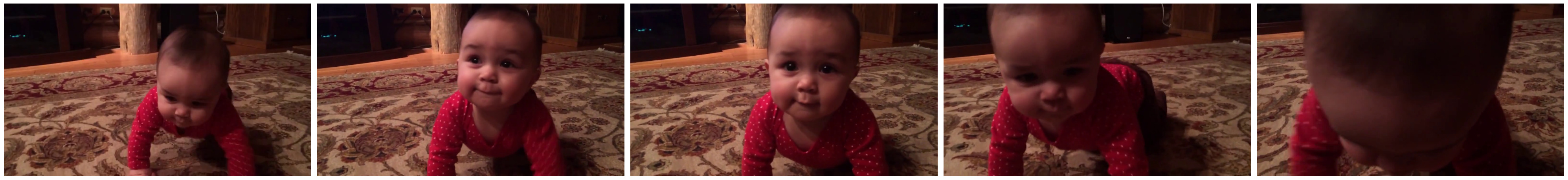}
    \vspace*{-5mm}
   \caption*{a baby crawling on the floor and looking at the camera}
    \end{subfigure}

    \begin{subfigure}{\textwidth}
    \includegraphics[width=\textwidth]{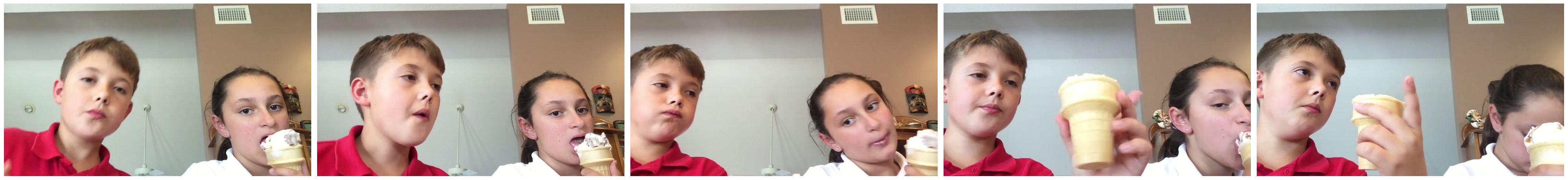}
    \vspace*{-5mm}
   \caption*{a boy and a girl eating ice cream}
    \end{subfigure} 

    \begin{subfigure}{\textwidth}
    \includegraphics[width=\textwidth]{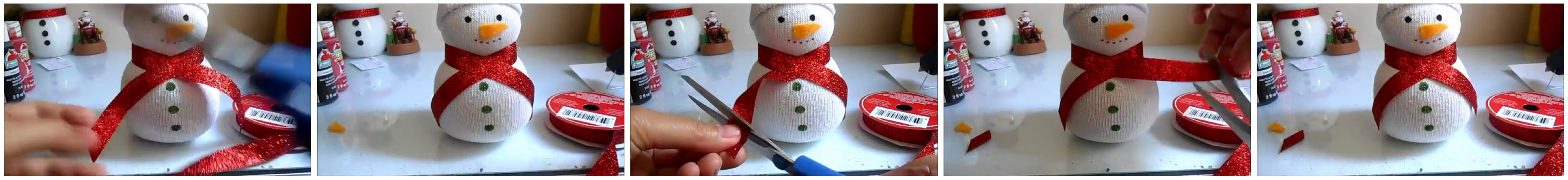}
    \vspace*{-5mm}
   \caption*{a video of how to make a snowman out of socks}
    \end{subfigure} 
    
    \caption{\textbf{Qualitative results:} Each row shows 5 frames sampled equidistantly from the video. Below the frames we print out the predictions made by our model.}
    \label{fig:qualitative_results}
\end{figure}

\begin{figure}
\captionsetup{justification=centering}
    \centering
    
    \begin{subfigure}{\textwidth}
    \centering
    \includegraphics[width=\textwidth]{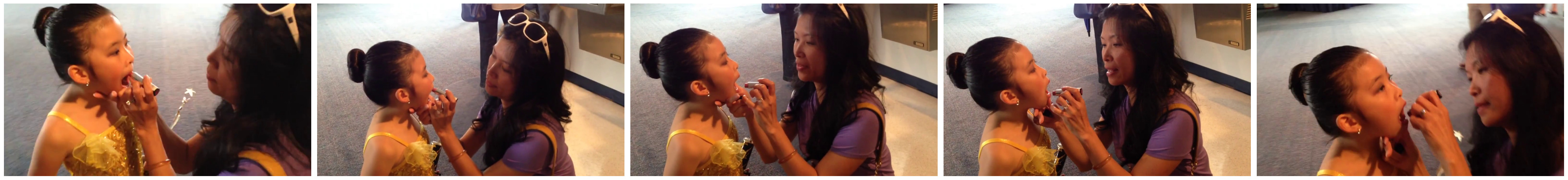}
    \vspace*{-5mm}
   \caption*{a little girl eating a piece of chocolate \\\textit{Failure case: Miss detecting the object (chocolate vs lipstick).}}
    \end{subfigure}    
    
    \begin{subfigure}{\textwidth}
    \centering
    \includegraphics[width=\textwidth]{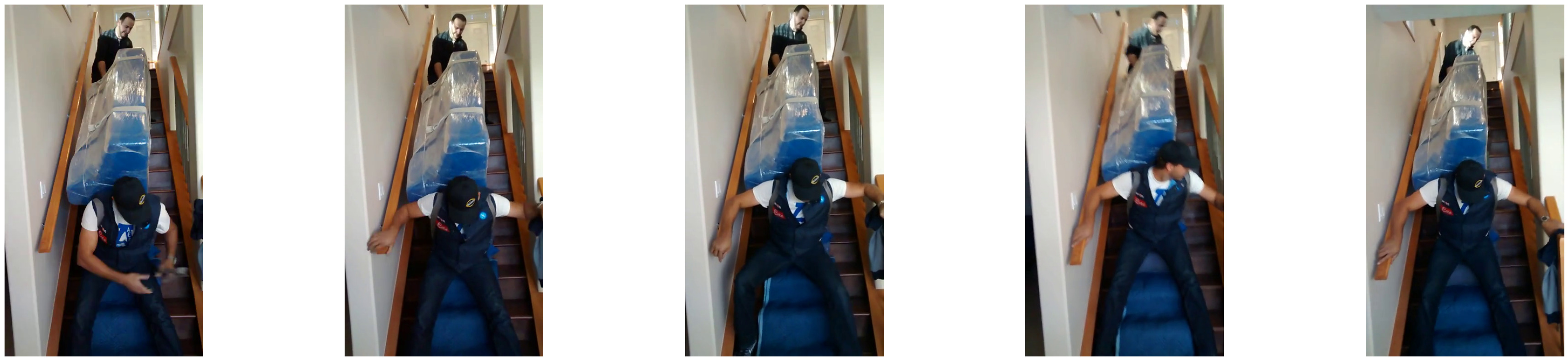}
    \vspace*{-5mm}
   \caption*{a man moving a mattress down a flight of stairs \\ \textit{Failure case: Wrong temporal action flow (here the mattress is moving up).}}
    \end{subfigure}    

    \begin{subfigure}{\textwidth}
    \centering
    \includegraphics[width=\textwidth]{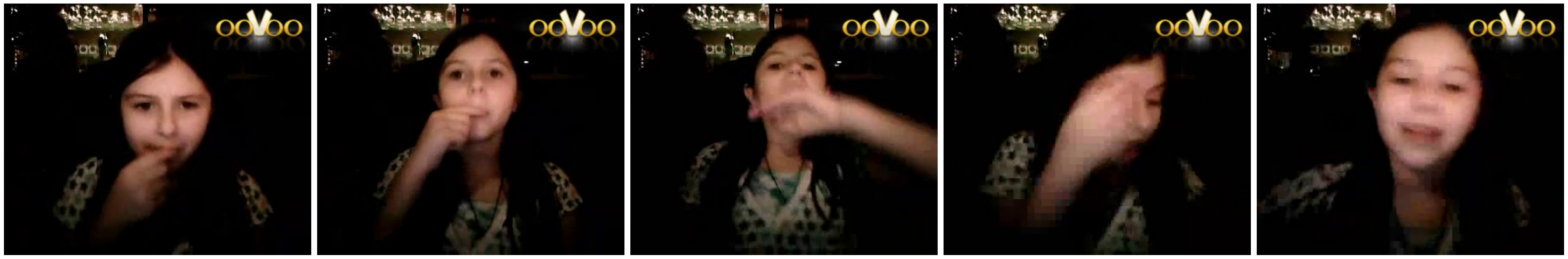}
    \vspace*{-5mm}
   \caption*{a little girl talking to the camera with the words subscribe on it \newline \textit{Failure case: Bias from the pre-trained model (the words subscribe on it).}} 
    \end{subfigure}    

    \begin{subfigure}{\textwidth}
    \includegraphics[width=\textwidth]{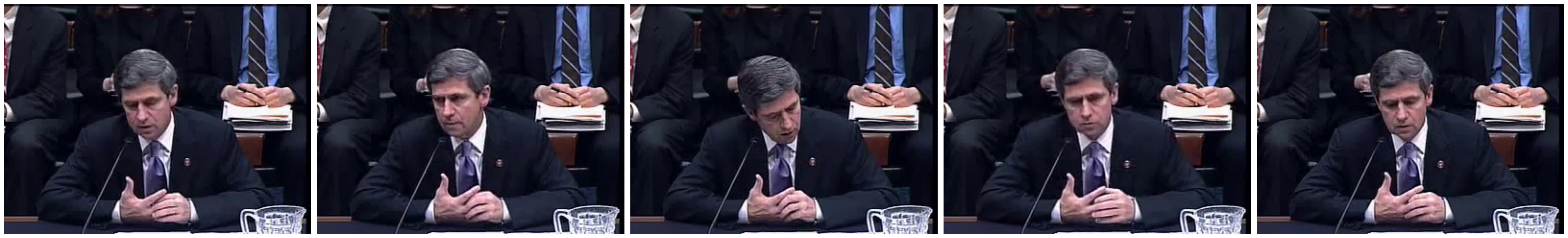}
    \vspace*{-5mm}
   \caption*{the president of the united states speaking at a senate committee meeting \newline \textit{Failure case: Hallucinating context-based details (i.e. that the person is the president of US). } } 
    \end{subfigure}    

    \begin{subfigure}{\textwidth}
    \includegraphics[width=\textwidth]{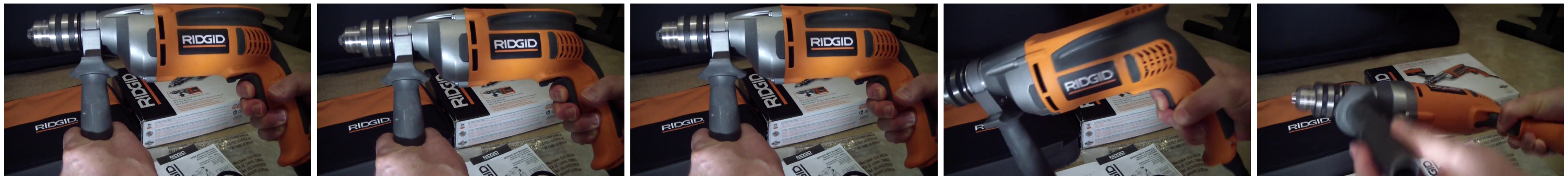}
    \vspace*{-5mm}
   \caption*{a person using a drill to drill a hole in a piece of wood \\ \textit{Failure case: Hallucinating plausible actions not shown in the video.}}
    \end{subfigure} 

\captionsetup{justification=justified}
    \caption{\textbf{Failure cases}. Each row shows 5 frames sampled equidistantly from the video. Below each set of frames we print out the predictions made by our model and a possible explanation associated to the particular failure case.}
    \label{fig:qualitative_results_failure_cases}
\end{figure}

\begin{figure}
\captionsetup{justification=centering}
    \centering
    
    \begin{subfigure}{\textwidth}
    \centering
    \includegraphics[width=\textwidth]{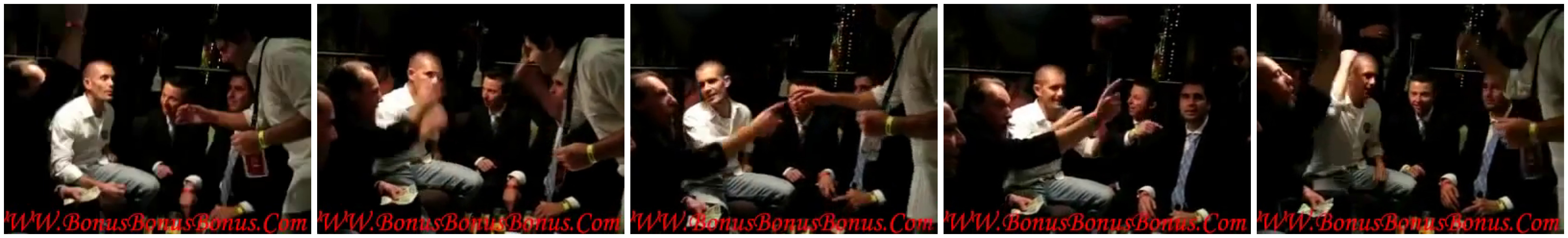}
    \vspace*{-5mm}
   \caption*{Step I: ``a group of people standing around a table with a lit candle''\\ Step II: ``a group of people standing around a table with a lit candle'' \\Step III: ``a man and a woman playing a game of rock paper scissors''\\Ground truth action (not used): ``rock scissors paper''}
    \end{subfigure}    
    
    \begin{subfigure}{\textwidth}
    \centering
    \includegraphics[width=\textwidth]{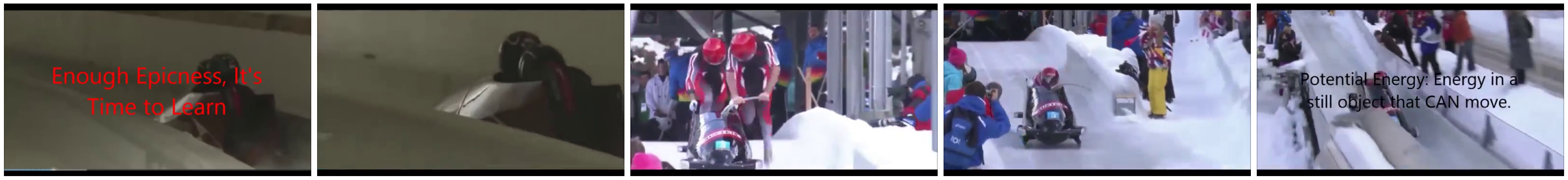}
    \vspace*{-5mm}
   \caption*{Step I: ``a snowboarder falling off the side of a building''\\ Step II: ``a bobsled bobsled race'' \\Step III: ``a bobsleigh race''\\Ground truth action (not used): ``bobsledding''}
    \end{subfigure}    

    \begin{subfigure}{\textwidth}
    \centering
    \includegraphics[width=\textwidth]{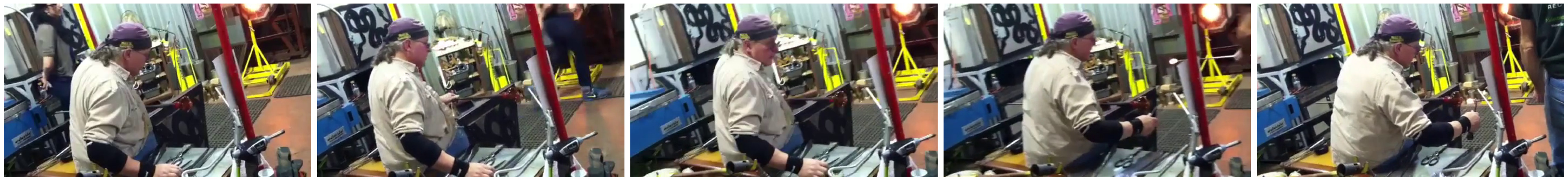}
    \vspace*{-5mm}
   \caption*{Step I: ``a man in a hat and sunglasses working on a piece of metal''\\ Step II: ``a glass blower blowing molten into the glass - blower stock videos and b - roll footage'' \\Step III: ``a glass blower blowing glass''\\Ground truth action (not used): ``blowing glass''}
    \end{subfigure}    

\captionsetup{justification=justified}
    \caption{\textbf{Qualitative results} showcasing the evolution of the most relevant pseudo-caption for each video after each retrieval step. }
    \label{fig:qualitative_results_progress}
\end{figure}

\end{document}